\documentclass{article}
\usepackage[utf8]{inputenc}

\usepackage[a4paper, margin = 1in]{geometry}
\usepackage{amsmath}
\usepackage{amssymb}
\usepackage{bm}
\usepackage{physics}\usepackage{makecell}
\usepackage{faktor}
\usepackage{enumerate}
\usepackage{amsthm}
\usepackage{empheq}
\usepackage{braket}
\usepackage{csvsimple}
\usepackage{authblk}
\usepackage{hyperref}
\usepackage[numbers,sort&compress]{natbib}

\begin{document}
\title{\textbf{Optimising Trotter-Suzuki Decompositions for Quantum Simulation Using Evolutionary Strategies}}

 \author[1]{Benjamin D.M. Jones}
  \author[1]{George O. O'Brien}
 \author[1]{David R. White}
 \author[2]{Earl T. Campbell}
 \author[1]{John A. Clark}

\affil[1]{Department of Computer Science, University of Sheffield}
\affil[2]{Department of Physics, University of Sheffield}

\maketitle

\begin{abstract}
One of the most promising applications of near-term quantum computing is the simulation of quantum systems, a classically intractable task. Quantum simulation requires computationally expensive matrix exponentiation; Trotter-Suzuki decomposition of this exponentiation enables efficient simulation to a desired accuracy on a quantum computer. We apply the Covariance Matrix Adaptation Evolutionary Strategy (CMA-ES) algorithm to optimise the Trotter-Suzuki decompositions of a canonical quantum system, the Heisenberg Chain; we reduce simulation error by around 60\%. We introduce this  problem to the computational search community, show that an evolutionary optimisation approach is robust across runs and problem instances, and find that optimisation results generalise to the simulation of larger systems.
\end{abstract}

\maketitle

\section{Introduction}

Practical Quantum Computers are likely to demonstrate speedup over classical computers on certain problems in the near future \cite{childs2017, childs2018, berry2007efficient, wiebe2011simulating, jones2018quest}.  Identifying applications that can exploit this advantage is of great importance; one of the most promising is simulating the time-evolution of quantum mechanical systems \cite{childs2017, childs2018, brown2010, berry2007efficient, berry2017exponential, somma2016, wiebe2011simulating, georgescu2014}. Simulating large systems involves expensive matrix calculations that are classically intractable, but can be performed much more efficiently with quantum computers.

One of the most exciting applications of quantum simulation is the modelling of complex molecular bonds and reactions. This facilitates the discovery of  lower-energy pathways for chemical reactions and the design of new catalysts, holding great potential for impact in medical research. Simulation of quantum chemistry may also play a significant part in the development of future high performance batteries. These and other simulations are beyond the reach of classical supercomputers \cite{hempel2018quantum, babbush2015, mcardle2018, luitz2015many, nandkishore2015many}.

The  ``quantum advantage'' that quantum computers offer on such problems is provided by the basic unit of quantum computing: the qubit. A qubit represents a single quantum state, and is implemented by the quantum properties of a physical system, such as the spin of an electron. Much like a classical bit, a qubit has two possible states: 0 or 1, denoted $\ket{0}$ and $\ket{1}$ in Dirac notation. However, a qubit can also exist in a \emph{combination} of these two states called a superposition, where the state of the qubit $\ket{\psi}$ is described by:

\begin{equation}
    \ket{\psi} = \alpha\ket{0} + \beta \ket{1}
\end{equation}

Here, $\alpha$ and $\beta$ are complex numbers termed \emph{probability amplitudes}\footnote{\emph{Why} nature should behave in this way is left as an exercise for the reader: to quote Feynman: ``we call [$\alpha$] a probability amplitude, because we do not know what it means.''}. Although the qubit is quite literally in a simultaneous combination of two states, under measurement it can only be observed as residing in one of them; the likelihood of the qubit being found in either state when measured is given by the square of the magnitude of the coefficients: $p(0) = |\alpha|^2$ and $p(1) = |\beta|^2$.

As these probabilities must sum to $1$, the state of a single qubit can be represented as a vector of unit length and all possible states of a single qubit can be represented as the surface of a sphere. The quantum analogues of logic gates are thus unitary matrices, as they must preserve the length (norm) of a qubit; they rotate the qubit, moving its state from one point on the surface of the unit sphere to another. Operators acting on more than one qubit work in an analogous manner in higher dimensional state spaces.

In practice, quantum computing hardware consists of a small set of physically implementable unitary gates. In almost all models, each gate is applied to a single qubit or a pair of qubits. If a problem can be broken down into operations that map to these gates, it is tractable on a quantum machine. As these gates reflect unitary matrices operators, simulating a quantum mechanical system on a quantum computer is therefore tractable if we can efficiently decompose the system into a product of such unitary matrix operators, each acting on small subsets of the total system. We achieve this by mapping each quantum property of the simulated system to a single logical qubit on the quantum computer, and changes in the quantum property of a system are simulated by operations on one or more qubits: even near-term quantum machines with a modest number ($n>40$) of qubits will surpass the ability of classical computers to simulate quantum systems. \cite{jones2018quest, preskill2018, brown2010}.

In order to implement a simulation on a quantum computer, a quantum program or `circuit' must be designed, breaking down the simulation into discrete steps. There are many methods of decomposition, including the popular Trotter-Suzuki method \cite{childs2017, berry2007efficient, hatano2005finding, dhand2014} used in this paper. Given a model of a system (such as its Hamiltonian, which is a matrix describing its energy), and a desired accuracy and time-scale, a sequence of operations i.e. a quantum circuit is derived. In this paper, we investigate the use of evolutionary computation to further optimise Trotter-Suzuki decompositions, by reducing simulation error or lowering the number of operations (the ``gate count'') required.

\section{Background}
We now give an overview of the mathematics describing quantum simulation, and the decompositions used to produce a circuit that will simulate the system on a quantum computer. We include only the minimum required to follow our work; see \cite{georgescu2014,shende2006, brown2010} for a more detailed overview. For those interested in quantum computing in general, excellent introductions are provided by \cite{rieffel2000introduction, nielsen2002quantum}.

\subsection{Quantum Simulation}

The state of a quantum system at time $t$ is represented by a state vector $\ket{\psi (t)}$, and its dynamics are determined by a Hamiltonian $H$ (a Hermitian matrix). The system will evolve over time according to the Schrodinger equation\footnote{We set Plank's constant $\hslash =1$.}:

\begin{equation}
    i \pdv{}{t} \ket{\psi (t)} = H \ket{\psi (t)}
\end{equation}

If $H$ is time independent, the solution to the above is given by:

\begin{equation}
    \ket{\psi (t)} = \exp(-iHt) \ket{\psi{(t=0)}}
    \label{eq:FACTOR_OUT}
\end{equation}

That is, given the vector $\psi{(t=0)}$ representing the system at time $t=0$, we can calculate the state of the system at time $t$ by multiplying it by the exponential $e^{-iHt}$. This requires the exponentiation of the matrix $H$, which can be calculated to any desired order using a Taylor expansion: it is this exponentiation of large matrices that makes classical simulation so costly. Combining $-i$ and $t$ into a single coefficient, we denote this calculation as:

\begin{equation}
    U(\lambda) = \exp(\lambda H)\label{eqU}
\end{equation}

The operator $U(\lambda)$ acts upon the entire system, which makes it difficult to implement as a single hardware operation. Fortunately, many Hamiltonians of interest can be written as the sum of $L$ local matrices $H_j$, describing interactions and state transitions of smaller parts of the system:

\begin{equation}
    H = \sum_{j=1}^L H_j
\end{equation} 

We therefore must calculate:

\begin{equation}
    U(\lambda) = \exp(\lambda H) = \exp\bigg( \lambda \sum_{j=1}^L  H_j \bigg)
\end{equation}

This is not straightforward, however, as $H_j$ are matrices and therefore do not in general commute:

\begin{equation}
    \exp(A+B) \neq \exp(A)\exp(B)
\end{equation}

The implementation of simulation on a quantum computer concerns the decomposition of this exponential of a matrix sum into smaller operations that can be efficiently implemented as quantum gates, whilst allowing for the non-commutativity of these matrices. We wish to minimise the number of quantum gates required to approximate  Equation \eqref{eqU} to within some desired accuracy $\epsilon$. Given $H$, $\lambda$ and $\epsilon$, we seek to find a sequence of efficiently implementable gates $\lbrace P_k \rbrace^m _ {k=1}$ with minimal length (i.e. minimal $m$) such that

\begin{equation}
    \norm{\exp(\lambda H ) - \prod_{k=1}^m P_k} <  \epsilon
\end{equation}

The matrix norm $\norm{\cdot}$ returns the largest singular value; it describes the deviation between Equation \ref{eqU} and an approximation of it. This metric provides an appropriate indicator of the error between our target matrix and our approximation; it will be used in this paper as a cost function.

\subsection{Trotter-Suzuki Decomposition}
Trotter-Suzuki decomposition is concerned with the efficient approximation of matrix exponentiation; it is applicable to matrix exponentiation in general, and has applications outside of quantum simulation. In this section we outline the foundational results presented in terms of asymptotic upper bounds: for the lower orders the reader is encouraged to convince themselves of their validity, for the general case we refer to \cite{suzuki1991, suzuki1990} and related work.

Given a matrix exponential over a decomposable matrix as in Equation \ref{eqU}, the Suzuki formula $S_n$ gives the $n$th order approximation to the exponential, i.e. it is equal to the Taylor expansion of that exponential with some higher-order error terms. For example:

\begin{align}
S_1(\lambda) &:=  \prod_{j=1}^L e^{\lambda H_j} = U(\lambda) + \mathcal{O}(\lambda^2) 
\end{align}

The first order Suzuki decomposition $S_1$ is the product of the exponential of each individual $\lambda H_j$, and is equal to the desired exponential $U(\lambda)$ plus some error term of order $\lambda^2$. Crucially, the individual  $e^{\lambda H_j}$ calculations can often be implemented as single-operation rotations of qubits on a quantum computer, enabling efficient simulation.

The second order decomposition can be given using the `trick' of multiplying the exponentials in forward and reverse order, reducing $\lambda$ appropriately:

\begin{align}
S_2(\lambda) &:=  \prod_{j=1}^L e^{\frac{\lambda}{2} H_j}\prod_{k=L}^1 e^{\frac{\lambda}{2} H_k} = U(\lambda) + \mathcal{O}(\lambda^3)
\label{eq:suzuki_second_order}
\end{align}

This decomposition compensates for non-commutativity and eliminates some error terms; its effect can be confirmed by examining the second order of the Taylor expansion.
In general, the $S_{2k}$th order decomposition is given by the recurrence relation:

\begin{align}
S_{2k}(\lambda) &:= S_{2k-2}(p_k \lambda)^2 S_{2k-2}((1-4 p_k) \lambda) S_{2k-2}(p_k \lambda)^2 \nonumber\\ 
&= U(\lambda) +\mathcal{O}(\lambda^{2k+1})
\label{eq:suzuki_recurrence}
\end{align}

Suzuki proves in \cite{suzuki1991} that $S_{2k}= S_{2k-1}$, and hence only even orders are considered. Crucially, this recurrence relation introduces the value $p_k$ used to define the coefficients. Suzuki specifies values for $p_k$ as follows:

\begin{equation}
    p_k:= (4 - 4^{\frac{1}{2k-1}})^{-1} 
    \label{eq:p_vals}
\end{equation}

Equations \ref{eq:suzuki_recurrence} and \ref{eq:p_vals} arise from a set of more general equations \cite{suzuki1991, ruth1983canonical}; they are mathematically designed to cancel relevant orders in the Taylor expansion. We observe that for a concrete problem adherence to these formulas is not crucial (as our only metric is error reduction), which motivates our search.

We can then apply the Trotter formula (also known as the Lie product formula) to further reduce our error. Intuitively, this takes the Suzuki approximation and `timeslices' it over $r$ steps:

\begin{equation}
\exp( \lambda \sum_{j=1} ^L H_j ) = S_{2k} \bigg ( \frac{\lambda}{r} \bigg )^r + \mathcal{O}\bigg ( \frac{\lambda ^{2k+1}}{r^{2k}} \bigg )  \hspace{30pt} k \in \mathbb{N}
\label{eq:time_slice}
\end{equation}

This general result is again taken from \cite{suzuki1991, suzuki1990}, and ensures that our error tends to zero as $r$ tends to infinity. It is important to note that we have two parameters that determine our approximation method: $k$, controlling the order of the Suzuki approximation, and $r$. There is some tension between these parameters: for a given gate count, we can use extra gates to adopt a higher-order Suzuki approximation, or alternatively to time-slice more finely; there are complicated trade-offs involved.


Any higher order approximation is ultimately given by this recursion, resulting in a sequence of second order approximations, each parameterised by a phase (the coefficient in the exponent, e.g. $\alpha$ in $e^{\alpha X}$). The phases are determined by our original $\lambda$ and the $p_k$ values in Equation \ref{eq:suzuki_recurrence}. Hence any approximation of the form $S_{2k} \Big ( \frac{\lambda}{r} \Big )^r$ can be expanded to:

\begin{equation}
    S_{2k} \bigg ( \frac{\lambda}{r} \bigg )^r = S_2~(\lambda_1)~S_2(\lambda_2) ... ~S_2(\lambda_M)
\end{equation}

Where $M = r \times 5^{k-1}$, and the total number of gates (exponentials) is $2L \times r \times 5^{k-1}$. Thus an approximation for a given order $2k$ and a given `timeslice' $r$ is fully specified by a vector of phases, where the components sum to our original $\lambda$:

\begin{equation}
    \bm{\lambda} = \begin{pmatrix}
    \lambda_1 \\
    \vdots \\
    \lambda_M
    \end{pmatrix}
\end{equation}

\begin{equation}
    \sum_{j=1}^M \lambda_j = \lambda
\end{equation}

For example, in the case of $S_{4} \bigg ( \frac{i}{3} \bigg )^3 $ $(k=2, r=3, \lambda=i)$:

\begin{multline}
    \bm{\lambda} = \bigg ( \frac{i p_2}{3},\frac{i p_2}{3},\frac{i (1-4 p_2)}{3},\frac{i p_2}{3},\frac{i p_2}{3},\frac{i p_2}{3},\frac{i p_2}{3},\frac{i (1-4 p_2)}{3} , \frac{i p_2}{3},\frac{i p_2}{3},\frac{i p_2}{3},\frac{i p_2}{3},\frac{i (1-4 p_2)}{3},\frac{i p_2}{3},\frac{i p_2}{3} \bigg )^T
    \label{eq:phases}
\end{multline}

The values given by Suzuki are only one possible choice, and the search for alternatives that may reduce simulation error is the focus of this paper. The vector of these $p$ values and the larger vector of $\lambda$ values define search spaces over which to apply search methods.

\subsection{Case Study: The Heisenberg Chain}
Simulation of the Heisenberg chain is a standard benchmark often used in the context of quantum simulation \cite{childs2017, childs2018, luitz2015many, nandkishore2015many,pal2010many}, and we use this example system in our experimentation. It describes a one-dimensional `chain' of particles (actually a loop) and the pairwise interactions of adjacent particles.

The Hamiltonian that describes our Heisenberg chain is:

\begin{equation} \label{hchain}
H = \sum^n _{j=1} \bigg [X_j X_{j+1} + Y_j Y_{j+1} + Z_j Z_{j+1} + v_j Z_j \bigg ] 
\end{equation}

This is the Hamiltonian governing a system containing $n$ qubits. The $j$th and $(j+1)$th qubits interact, where indices are modulo $n$ so that the last qubit interacts with the first. 

An instance of this Heisenberg chain is defined by the Hamiltonian, the number of qubits $n$, and a vector $v$ of the values $v_j$ for each qubit in the system. We allow for disorder in the physical system that makes $v$ a random variable. For our purposes, $v_j$ is a scalar drawn uniformly random within the interval $[-1,1]$, as also considered in \cite{childs2017, childs2018, pal2010many}. A simulation is then fully defined by (a) the description of the Heisenberg chain and (b) the period of time $t$ that we wish to simulate the evolution of that chain over. We observe that this Hamiltonian is decomposed into the sum of $4n$ local matrices $H_j$, enabling us to apply Suzuki decomposition to efficiently simulate the system.

The $H_j$ matrices are defined in terms of the well-known Pauli matrices:

\begin{equation}
X=\begin{bmatrix}
0 & 1 \\
1 & 0
\end{bmatrix} \hspace*{20pt} Y=\begin{bmatrix}
0 & -i \\
i & 0
\end{bmatrix} \hspace{20pt} Z=\begin{bmatrix}
1&0 \\
0&-1
\end{bmatrix}
\end{equation}

These matrices do not commute with one another. A Pauli matrix $P$ acting on the $j$-th qubit in a $n$-qubit system gives rise to the $P_j$ matrix via the Kronecker (tensor) product:

\begin{equation}
P_j : = I \otimes ... \otimes P \otimes ... \otimes I
\end{equation}
where $I$ is the identity and $P$ is in the $j$-th position, and $P_i$ and $Q_j$ commute unless  $i =  j$ and $P \neq Q$.

Further background on the Heisenberg Chain and related problems can be found in \cite{childs2017, childs2018, luitz2015many, nandkishore2015many, pal2010many}.

\section{Experimentation}
We have observed that the phase vectors produced by a Suzuki decomposition are not the only possible choices, and we therefore apply search to explore those vector spaces, using the Heisenberg Chain as a motivating example. We take a simulation specified by $n$, $v$, and $t$, and seek to optimise the corresponding Suzuki decomposition given by a particular $k$ and $r$, in order to reduce simulation error. The Suzuki decomposition will determine the gate count, which we do not modify. Solutions to this problem are real-valued phase vectors; we factor out the $-it$ in Equation \ref{eq:FACTOR_OUT}.

We report the error for the optimised vector compared to the original error for the Suzuki solution. We mainly consider problems in which $n=5$ and set $t=2n$; systems of five qubits are large enough to be non-trivial but small enough to be efficiently simulated for fitness evaluation. It has been remarked that the simulation time should scale linearly with the system size \cite{childs2017, childs2018}.

A related problem is to fix an error threshold (such as $\epsilon = 0.001$, again as considered in \cite{childs2017, childs2018}) and minimise the number of gates required. Note that minimising error at a fixed gate count can result in obtaining a lower gate count that surpasses a particular threshold, by reducing the error of an otherwise insufficient solution.

\subsection{Research Questions}
We pose two research questions:

\subsubsection*{RQ1: Can we consistently improve on Suzuki solutions?}

As the Suzuki decompositions are derived theoretically, and the choices of coefficients in the recursion relation are  designed to construct agreement at a given order, it is not clear a priori that \emph{any} improvement can be made by modifying phases. We apply search to optimise these values; to the best of our knowledge we are the first to attempt this optimisation, and so any improvement is of scientific interest. Further, we wish to know whether different problem instances (i.e. Heisenberg Chains with different values of $v$) can be optimised, and how robust these gains are across different instances. 


To be useful in practice, optimisation results must also be robust to the seed (across runs). Thus, we also examine the distribution of results across runs for the same problem instance.

\subsubsection*{RQ2: How well do optimised decompositions generalise?}
It is desirable for optimised phase vectors to generalise to other uses: in particular, it would be advantageous if optimisations generalise across different values of:

\begin{enumerate}
\item $v$, i.e. other problem instances of the same size
\item $n$, the number of qubits in the Heisenberg Chain
\item $t$, different simulation times
\item $r$, i.e different granularity of `time-slicing' in Equation \ref{eq:time_slice}
\end{enumerate}

Generalisation over $n$ or $t$ would allow us to first optimize for smaller circuits or shorter simulation time, and then apply the results of optimisation to larger or longer simulations. To assess generalisation, we evaluate the optimised phase vectors produced in answering RQ1 across different values of these parameters.

\subsection{Applying Search}

There are two candidates for optimisation: the vector of the coefficients in the Suzuki recurrence relation (specified by $k$), and the full phase vector finally produced by decomposition (specified by $r$ and $k$). In this paper, we primarily search over the Suzuki coefficient space, involving the time-slicing factor $r$ only in evaluation. We give an example of the form of this vector in equation \eqref{eq:p_vector}. The Suzuki coefficient space avoids some of the redundancy of the full phase vector, although it may sacrifice some of the potential for ``tailoring'' a simulation to a given solution.

\begin{figure}
\centering
\includegraphics[width=0.8\textwidth]{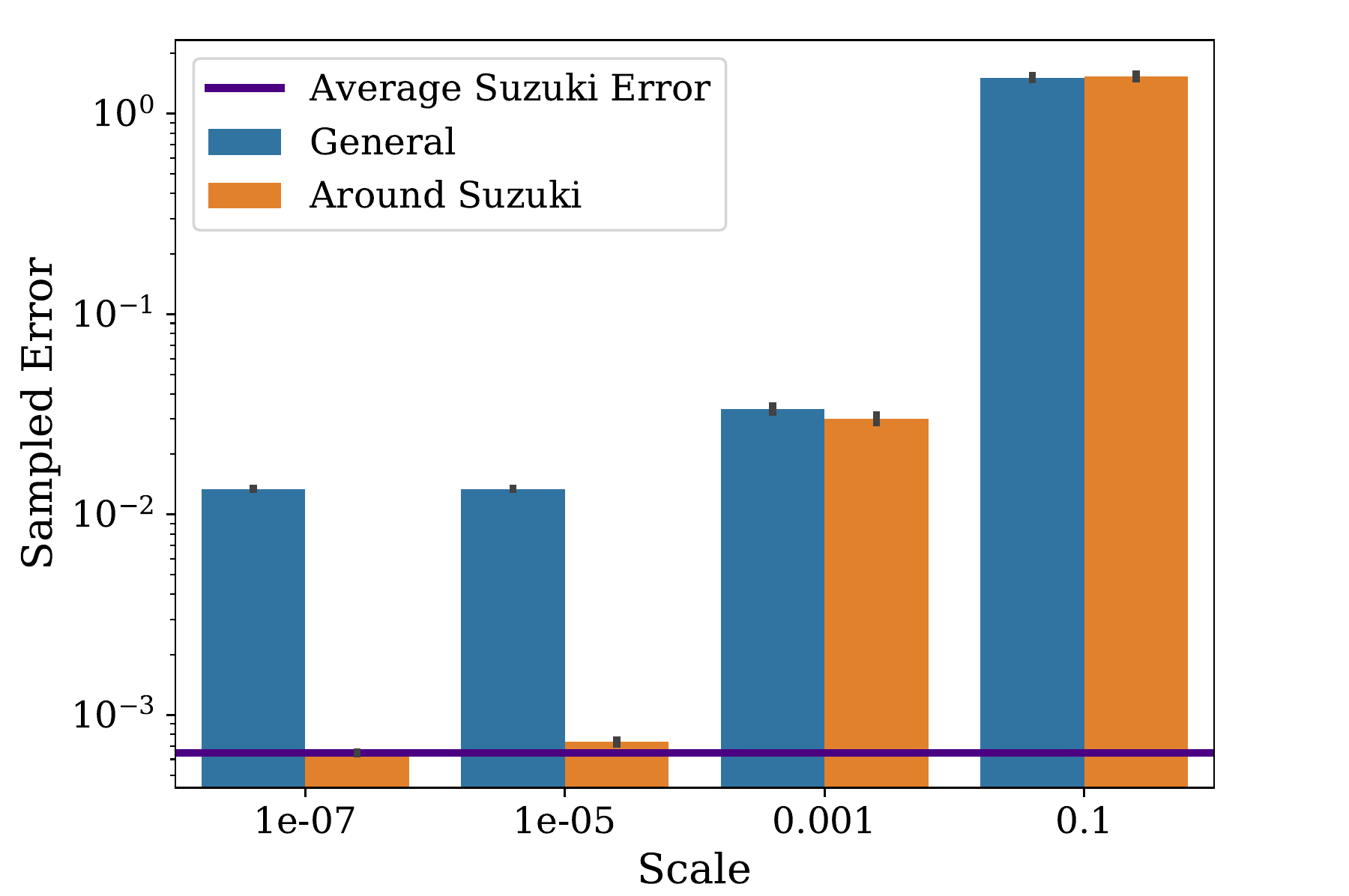}
\caption{Sampling the space of Suzuki coefficient vectors at different scales. Sampling was performed across the solution space, and close to the Suzuki solution. Most of the vector space contained solutions of very low quality. Based on this analysis, we started our search at the corresponding Suzuki solution. }\label{sampling}
\end{figure}

Random sampling and systematic hill-climbing of the solution space during pilot experiments revealed that most vectors are of fitness close to the worst possible value, with insufficient gradient to guide search to solutions even within a few orders of magnitude of the Suzuki error. We performed two types of sampling: the first drawing candidate $p$ vectors from a normal distribution centred on the inverse of the length of the vector, such that the expected sum of the vector is one\footnote{Observe that the case in which all $p$ values are equal to the inverse of the length is equivalent to the case $k=1$, with $r$ equal to the length of the vector.}. Explicitly, for $k=2$ this involves sampling values from a normal distribution centred on $0.2$. We performed this sampling for a range of standard deviations. The second type of sampling sampled a normal distribution with the Suzuki solution as its centroid. See Figure \ref{sampling} for an overview of our results: we looked at $10$ problems over $k=2$ and $k=3$ each over two relevant $r$ values, but only display data for $k=2$ and $r=125$. We drew $100$ samples from the space in each case, and the graph represents an average.

In light of these results, we take the Suzuki solution as the starting point for our search, concluding that the area around the Suzuki vectors is likely to be most fruitful.

Almost all our experiments apply CMA-ES \cite{Hansen2003} to the coefficient vector. We use CMA-ES because it represents state-of-the-art in evolutionary computation, and because pilot exploration of the search space revealed a highly multimodal landscape that necessitated a variable step size: CMA-ES adapts automatically. It was not possible to apply CMA-ES to the high-dimensional phase vector as the high number of fitness evaluations required was prohibitively expensive. Exploratory experiments indicated CMA-ES over the coefficient space greatly outperformed a simple ES on the full phase vector.

All the parameters of CMA-ES are left at their DEAP defaults where defaults are provided (these defaults are themselves well-justified: see \cite{Hansen2001}). We set the number of generations to 250, by which point all our pilot runs had converged, and we set the initial step size to $1 \times 10^{-7}/d$, where $d$ is the length of the vector being optimised (usually five). As the initial vectors sum to one, it is intuitive that the step size be inversely proportion to the vector length; the order of magnitude was chosen based on our initial sampling, and is  subsequently adapted by CMA-ES.
 
Optimising the Suzuki coefficients requires searching over a vector space of dimension $5(k-1)$. Our seed individual is a concatenation of vectors of the form $(p_k, p_k, 1-4p_k, p_k, p_k)$, from $k=2$ up to the desired value of $k$\footnote{Here we concatenate values, but we could also have considered expanding these coefficients via the recursion relation in Equation \ref{eq:suzuki_recurrence} to obtain a vector of length $5^{k-1}$. We choose to concatenate values to exploit structure in the solution space, and to reduce the search space size.}. For example, for the second-order decomposition our seed solution would be a vector of length five and the starting point for search (the centroid parameter to CMA-ES) is given by the coefficients in Equation \ref{eq:suzuki_recurrence}: 

\begin{equation}
\begin{pmatrix}
    p_2, & p_2, & 1-4p_2, & p_2, & p_2
\end{pmatrix}  \label{eq:p_vector}
\end{equation}
For $k=3$ we would evolve the following vector:
\begin{align}
    \bigg (\hspace{5pt} &p_2, \hspace{5pt} p_2, \hspace{5pt} 1-4p_2, \hspace{5pt} p_2, \hspace{5pt} p_2 ,
    \hspace{5pt} p_3, \hspace{5pt} p_3, \hspace{5pt} 1-4p_3, \hspace{5pt} p_3, \hspace{5pt} p_3\hspace{5pt} \bigg )
\end{align}

We refer to such vectors as `$p$-vectors'. The fitness or cost function $f: \mathbb{R}^M \longrightarrow \mathbb{R}$ then expands the vector according to the recursion relation in Equation \ref{eq:suzuki_recurrence} (for $k>2$) and includes a time-slicing factor $r$, as in Equation \ref{eq:time_slice}.

Explicitly for $k=2$, which is the case primarily considered here, the fitness function is specified by:

\begin{align}
    \bm{p} = (x_1 , ... , x_{M}) &\longrightarrow  \norm{\exp(-it H ) - \bigg [ \prod_{j=1}^{M} S_2 \bigg (\frac{-itx_j}{r} \bigg ) \bigg ]^r } 
\end{align}

It is somewhat counter-intuitive that to evaluate our approximation to a quantum simulation requires ``knowing the answer'', i.e. being able to efficiently calculate $\exp(-itH)$ on a classical computer! This paradox is resolved by noting that we calculate this exponentiation for small systems only; in particular, systems composed of low numbers of qubits (typically $n=5$ in our experimentation). We believe the optimisation of small simulations is still useful in gaining insight, especially if results can generalise to larger systems.

Both $p$-vectors and full phase vectors are constrained by Suzuki to sum to one, but we relax all constraints during search in order to obtain the lowest possible error: the constraints are theoretically-motivated rather than essential in practice. Relaxing the constraints also removes hard boundaries in the search space; the alternative of post-mutation normalisation is problematic, partly because it introduces non-linear behaviour when applying local search.

\subsection{Implementation}
We implemented a simple library describing Heisenberg chains, Hamiltonians, and Suzuki decompositions in Python, using the numpy \cite{oliphant2006guide} and associated libraries to perform the requisite linear algebra. Full code, experimental scripts, parameter files, analysis scripts, and results can be found online\footnote{\url{https://github.com/sheffieldquantum/qsim}}. We use the cachetools module in the standard Python library to reduce execution time, caching matrix exponentiation wherever possible.

To perform search, we rely on the DEAP evolutionary framework \cite{DEAP}, which implements standard ES and CMA-ES. We ran our experiments on a HPC cluster using Python libraries built with the Intel MKL to improve performance. A note on improving the efficiency of matrix exponentiation is given in Appendix \ref{app:efficient_exponentiation}.

\section{Results}
We find that for a given simulation (specified by $n$, $v$ and $t$) and Suzuki decomposition (specified by $k$ and $r$) we can achieve a significant reduction in simulation error. We demonstrate that our method is robust across many different problems (values of $n$, $v$, $k$ and $r$) and that optimised results also generalise: they achieve error improvements for other values of $v$ and $n$.

\subsection{RQ1: Reducing Simulation Error}
\label{sec:results_rq1}

\subsubsection*{Robustness across $v$ and multiple runs}
\begin{figure}
    \centering
    \includegraphics[width=0.7\textwidth]{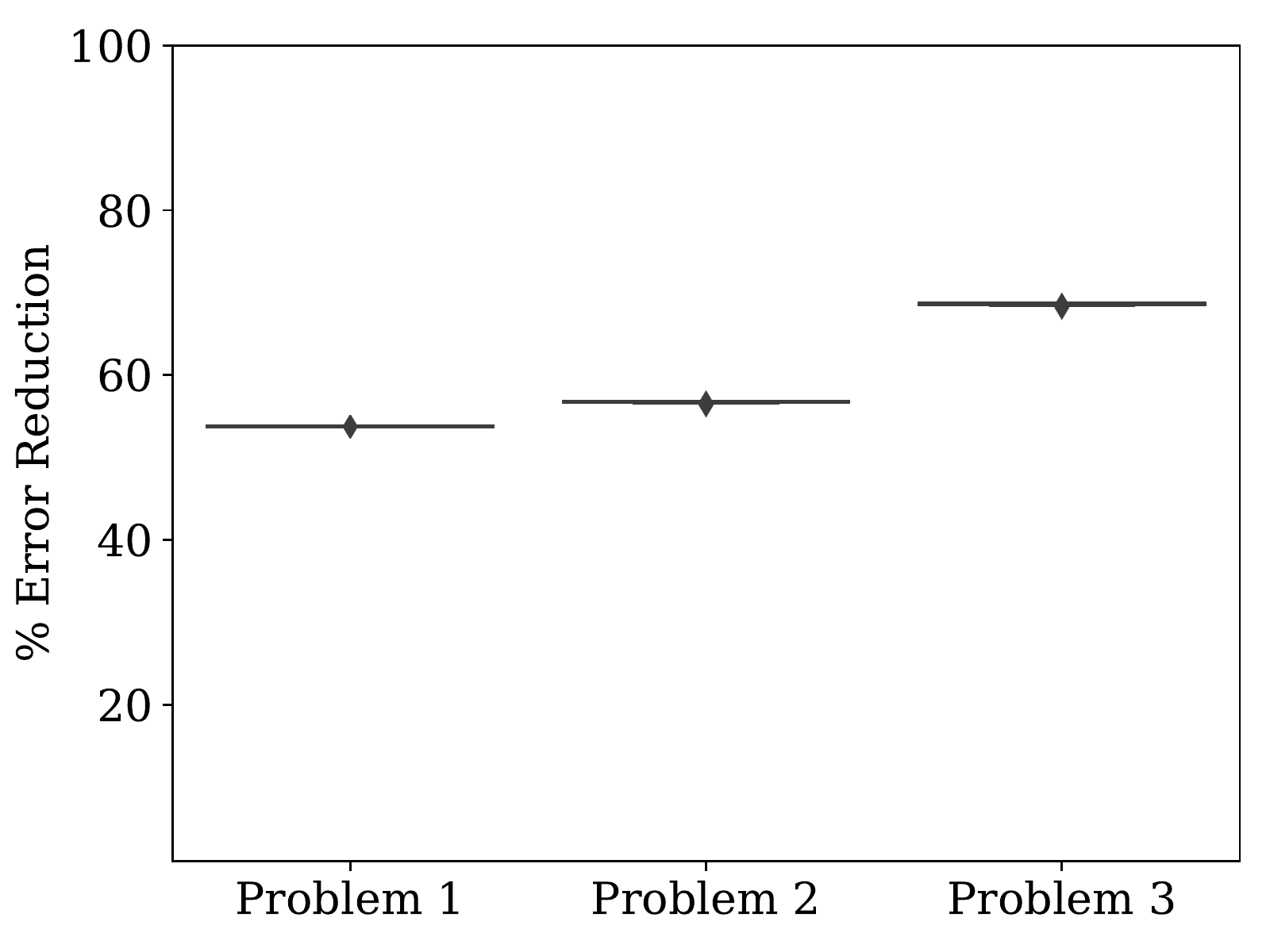}
    \caption{Boxplot of the error reductions for three Heisenberg Chains (three v vectors), each over 30 runs of CMA-ES. Here $n=5$, $k=2$ and $r=125$. The range in \% error reduction for each of the problems was approximately 0.071, 0.41 and 0.47 respectively. Variance is very low: CMA-ES performs consistently well.}
    \label{fig:boxplot_single_problem}
\end{figure}

Figure \ref{fig:boxplot_single_problem} shows the error reduction using CMA-ES on the Suzuki coefficient vector, for three problem instances over 30 runs. The median improvements over the three problems were $53.8\%$, $56.7\%$, and $69.6\%$ respectively, and similar error reductions were achieved for other problem instances

\begin{figure}
\centering
\includegraphics[width=0.8\textwidth]{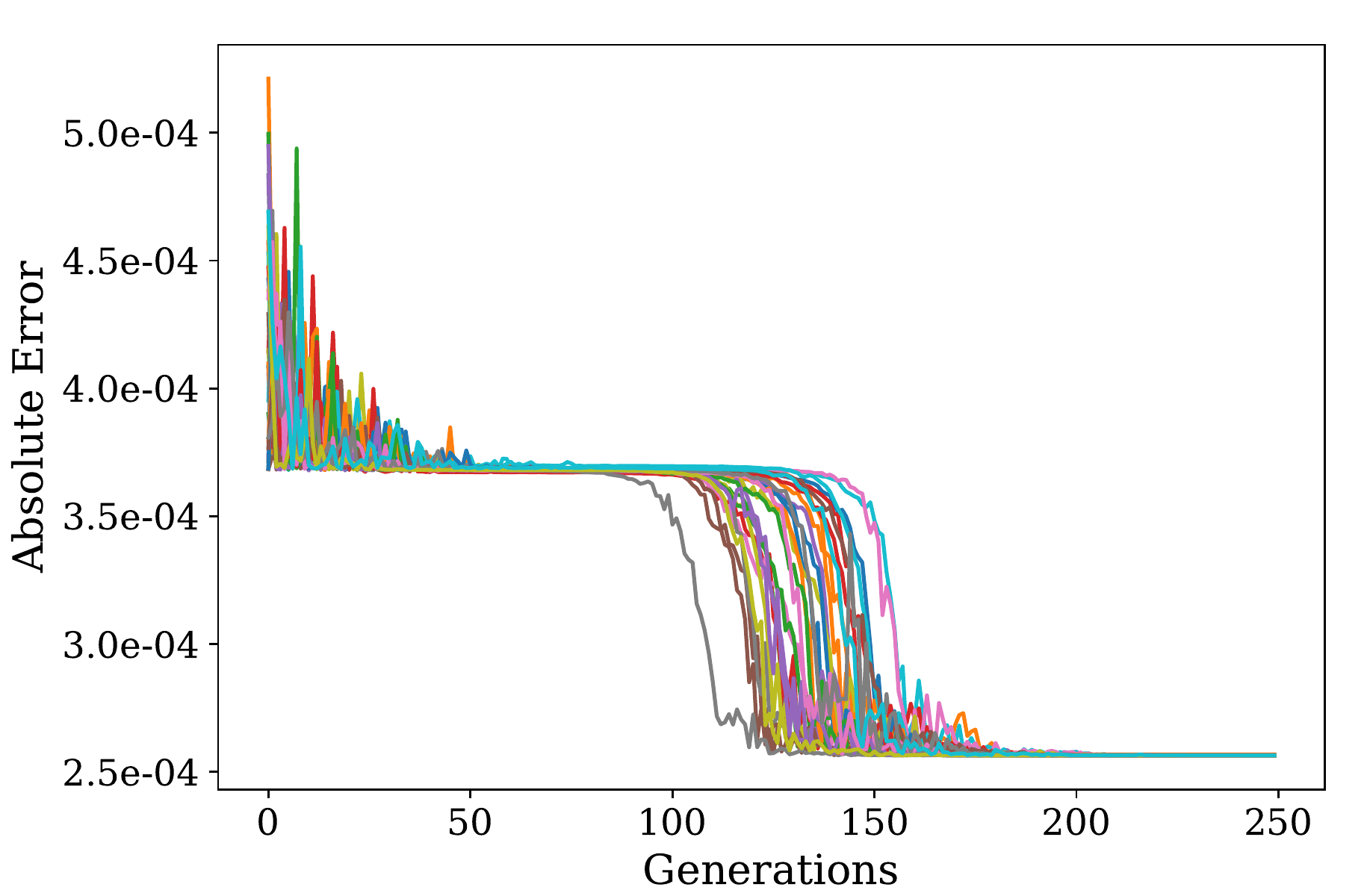}
\caption{Optimisation trajectory for CMA-ES over different runs for the same problem. The trajectories are very similar, reflecting the robustness of optimisation results across runs}\label{run plot}
\end{figure}

Figure \ref{run plot} shows multiple runs of CMA-ES on a given problem instance converging along similar trajectories; in pilot experiments we found greedy hill-climbing could often reach the initial plateau, but only CMA-ES was able to escape and make further progress. We conjecture that this is a result of its adaptive step size. The consistent behaviour suggests that different problems have similar landscapes, and that the search should be robust across problems.

\subsubsection*{Robustness across multiple problem instances}
\begin{figure}
\centering
\includegraphics[width=0.7\textwidth]{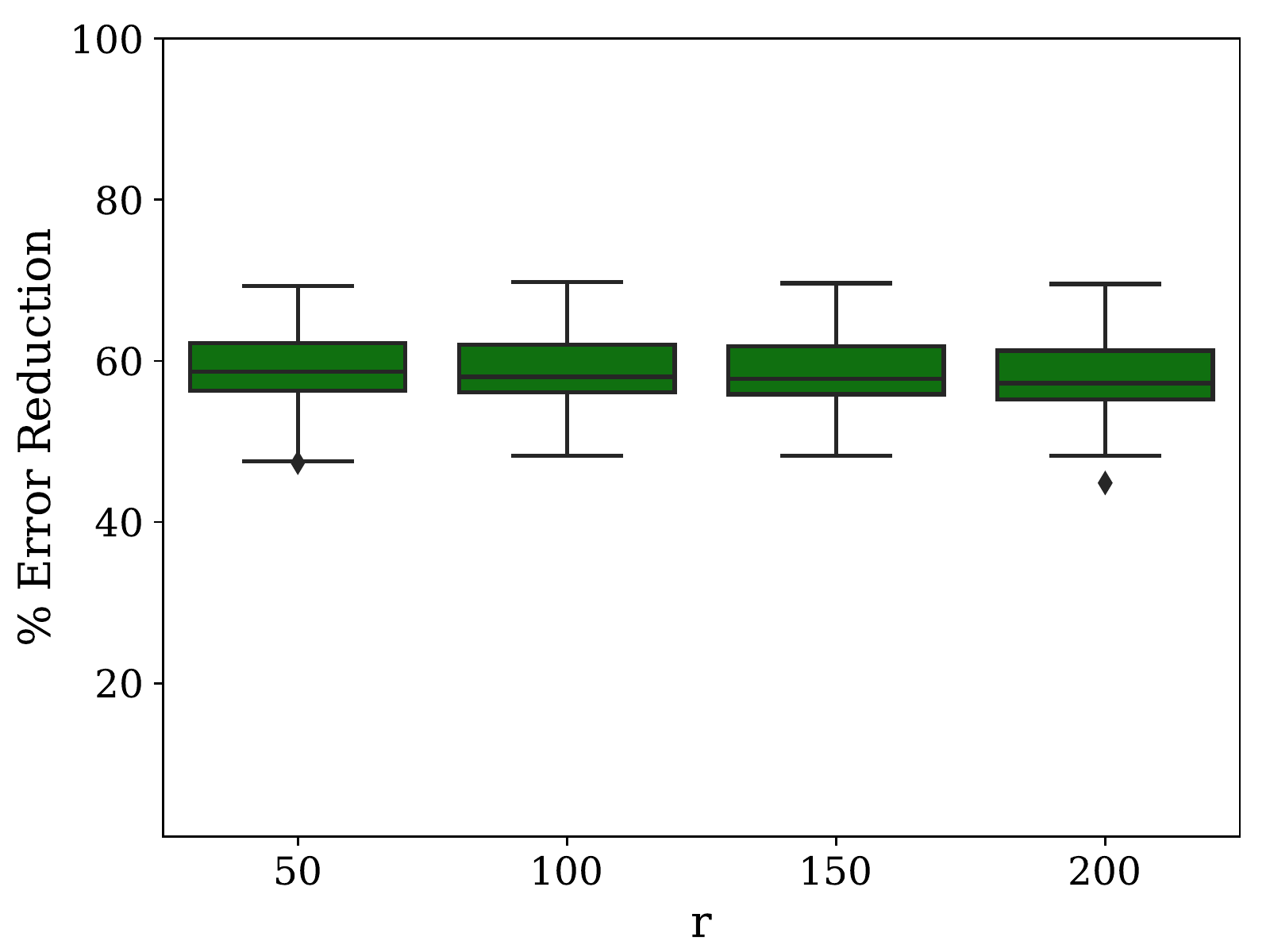}
\caption{Robustness of optimisation across different $r$; different levels of granularity of `time-slicing'. Error reduction is consistent across values of $r$.}\label{boxplot_r}
\end{figure}

Figure \ref{boxplot_r} displays error improvements across 30 different $v$ vectors at four different values of $r$. The reduction in error is consistent.

\subsubsection*{Robustness across gate count}

\begin{figure*}
\centering
\includegraphics[width=\textwidth]{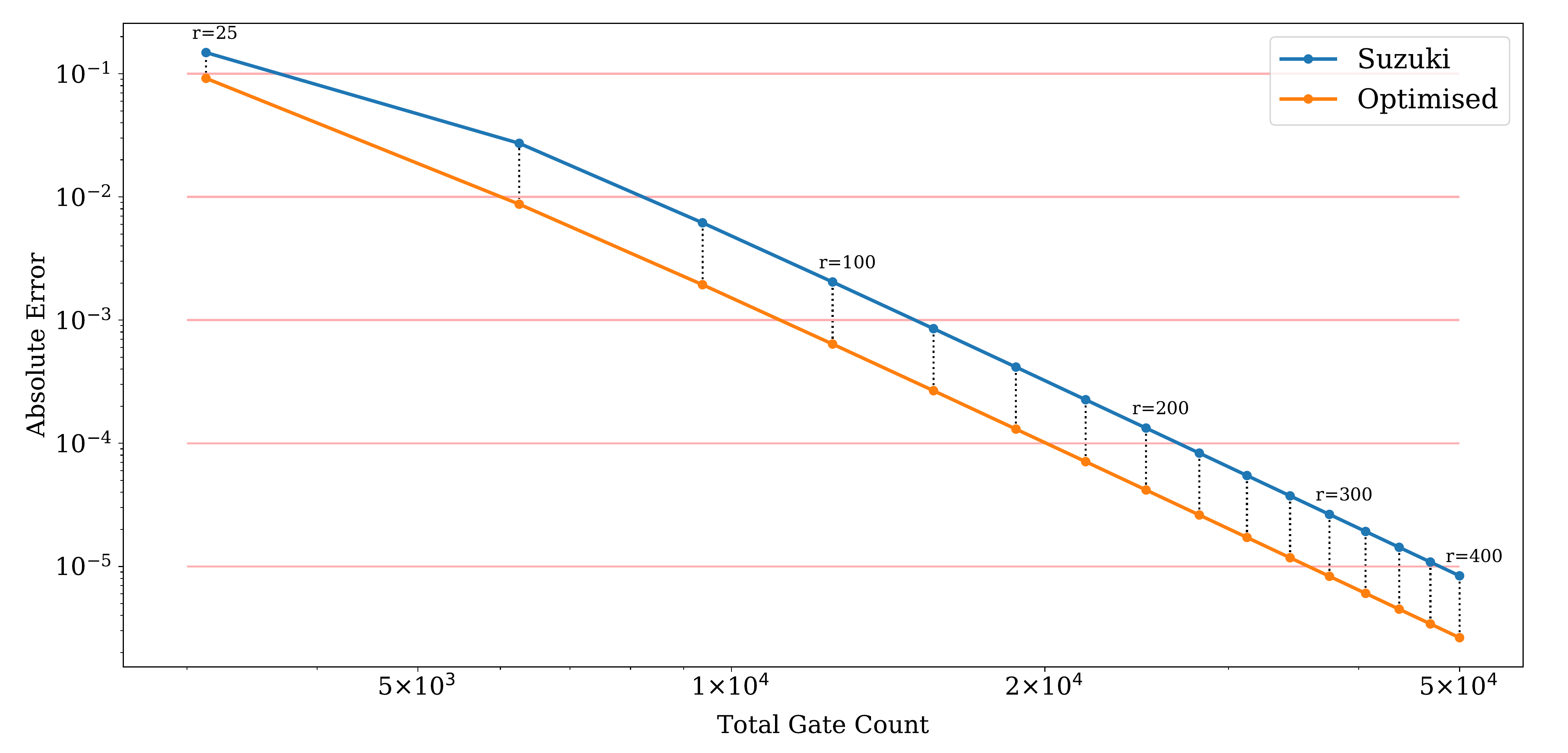}
\caption{Robustness of optimisation across total gate count, for n=5 and k=2. Total gate count is determined by $k$ and $r$. Here, $k$ is kept constant whilst $r$ is varied. Similar error improvements are found across different gate counts.}\label{r plot}
\end{figure*}

The gate count of a simulation is determined by the order of the Suzuki approximation used, and the `timeslicing' granularity given by $r$. Figure \ref{r plot} shows the results of optimising the second-order Suzuki approximation at different values of r for a specific Heisenberg chain. This result demonstrates that our approach can be applied at any desired value of $r$. Of particular note is the idea of `pulling solutions down' to provide error below a certain threshold (for example consider here $\epsilon=10^{-3}$) at a reduced gate count: we can obtain such error with $r=100$ as opposed to $r=125$. This demonstrates how improvements in error reduction can directly lead to improvements in gate count.

\subsubsection*{Robustness across n}
\begin{figure}
    \centering
    \includegraphics[width=0.7\textwidth]{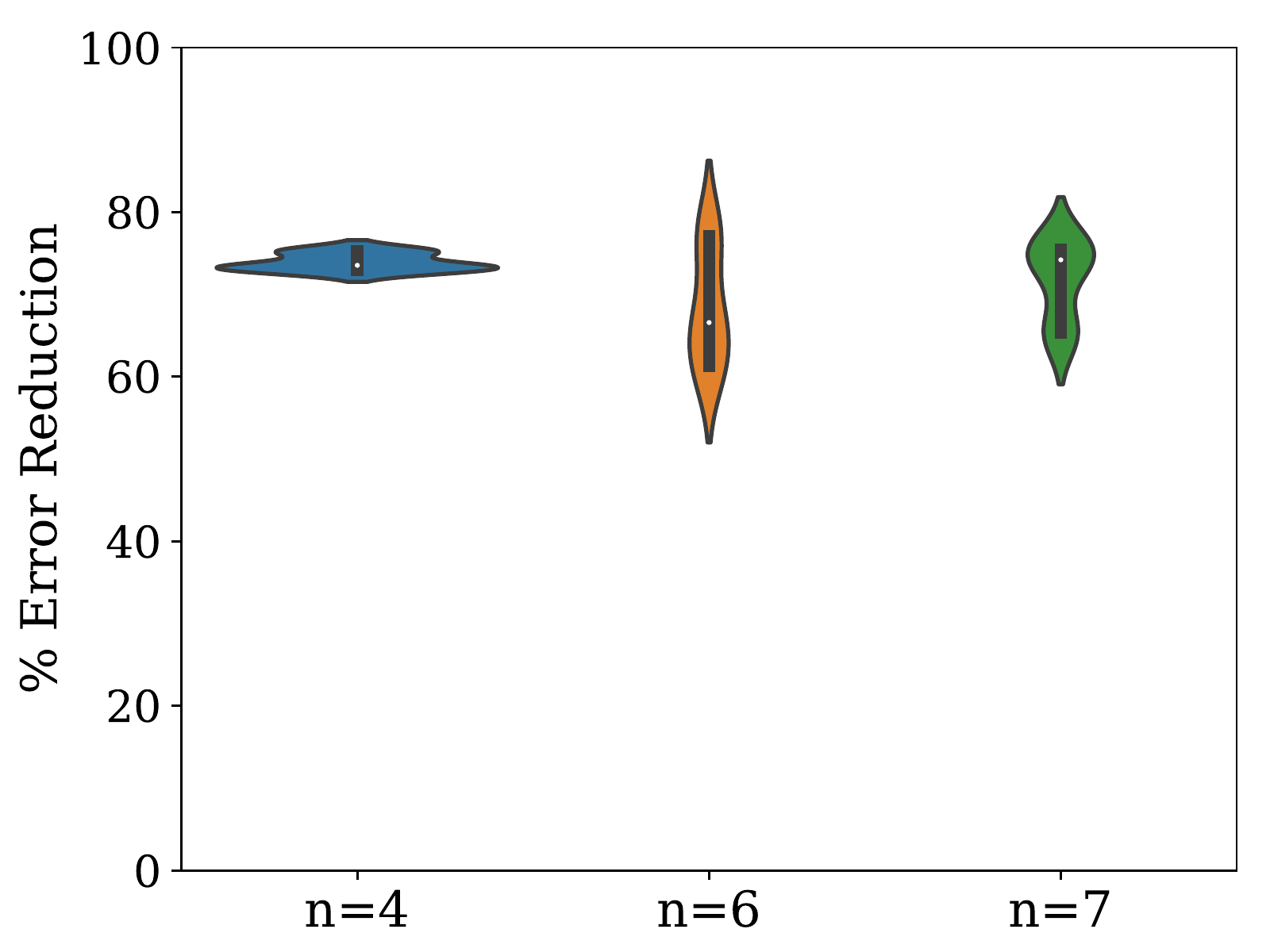}
    \caption{Reduction in error show robustness over systems composed by $n$ qubits.}
    \label{fig:robustness_n}
\end{figure}

Figure \ref{fig:robustness_n} shows that we can achieve similar error reductions for $n=4,6$ and $7$. For each $n$, the plot shows data over three problem instances and three seeds.

\subsubsection*{Robustness across k} 
\begin{figure}
\centering
\includegraphics[width=0.7\textwidth]{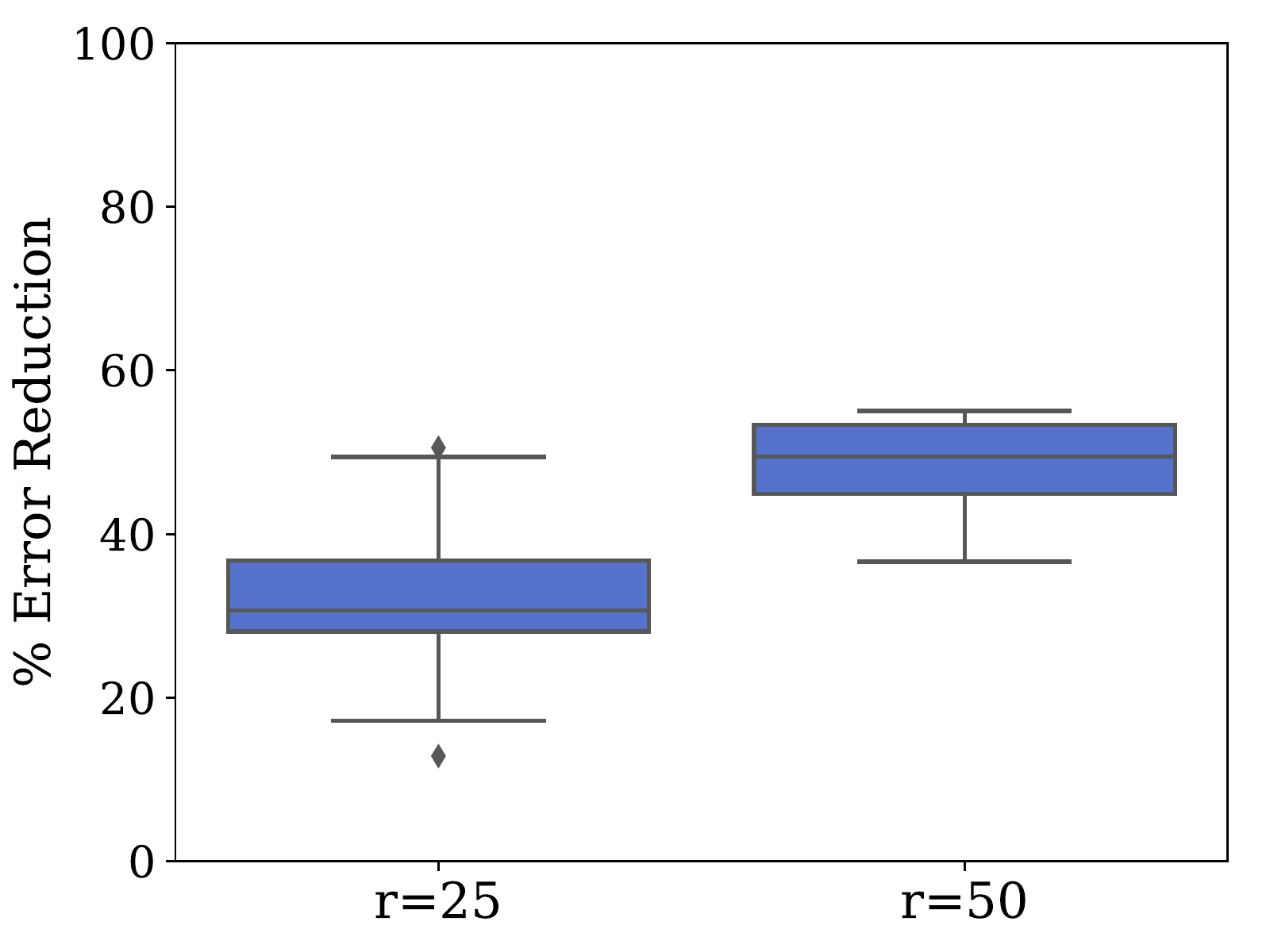}
\caption{Robustness of optimisation for k=3, across two values of $r$. Each boxplot summarises twelve v values and three seeds; as the resulting vector of coefficients is twice as large, these experiments were run for 500 generations.}\label{k=3 plot}
\end{figure}

Most of our results consider only $k=2$, i.e. the second-order Suzuki decomposition. Whilst this decomposition is often sufficient to efficiently achieve desired simulation error, we also report results for $k=3$ in Figure \ref{k=3 plot}. Results shown here are averages over $12$ problem instances, with three runs each. Our seed individual is a concatenation of relevant $p$ vectors at $k=2$ and $k=3$, explicitly it is a vector of length $10$:
\begin{align}
    \bigg (\hspace{5pt} &p_2, \hspace{5pt} p_2, \hspace{5pt} 1-4p_2, \hspace{5pt} p_2, \hspace{5pt} p_2 \hspace{5pt},
    &p_3, \hspace{5pt} p_3, \hspace{5pt} 1-4p_3, \hspace{5pt} p_3, \hspace{5pt} p_3\hspace{5pt} \bigg )
\end{align}

We achieve substantial improvement in error, although less than our results for $k=2$; even after running for 500 generations. We observe that error is continuing decrease after 500 generations have elapsed, and it is likely further runtime is required to compensate for the increase in the dimension of the search space.

\subsection{RQ2: Generalisation of Results}
We now measure how well results found in Section \ref{sec:results_rq1} generalise over other values of $v$ $r$, $n$ and $t$. Of greatest interest is the ability to generalise over $t$ and $n$, as any such behaviour would enable us to learn over the simulation of smaller systems or shorter simulation times and apply results to larger problems without requiring the ability to efficiently evaluate optimisations on those large systems.

\begin{figure}
    \centering
    \includegraphics[width=\textwidth]{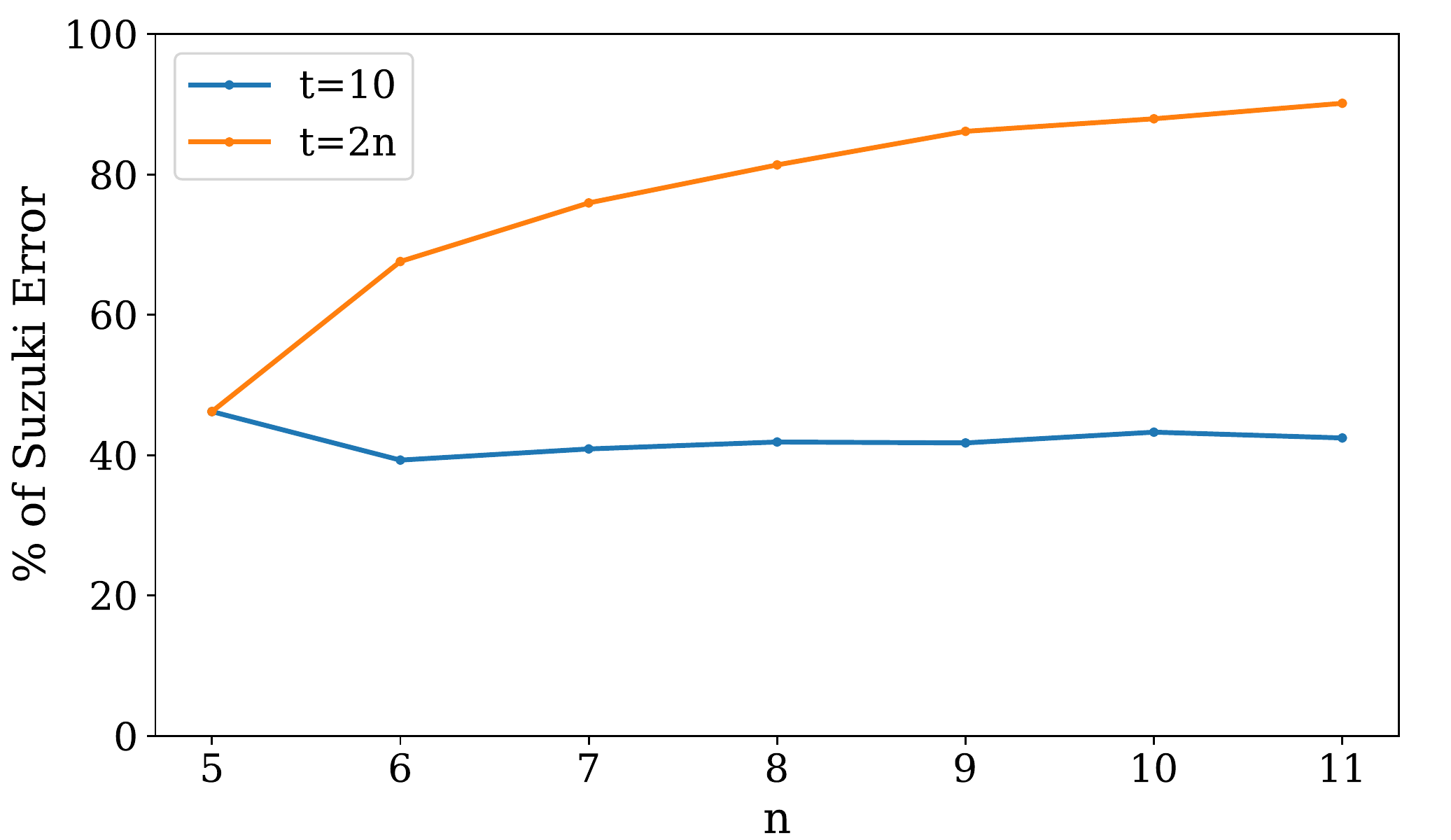}
    \caption{Generalisation over n and t, for a given optimised $p$-vector. Results generalise over $n$, suggesting optimisation results obtained on small systems can be applied to larger ones. No such generalisation is found over $t$, simulation time. Similar generalisation behaviour was found for other results.}
    \label{fig:generalisation_t_n}
\end{figure}

In Figure \ref{fig:generalisation_t_n} we see that generalisation over $t$ is poor, as simulation error grows with $t$ and eventually exceeds the error achieved with the original Suzuki solution. In contrast, we \emph{do} find that our results generalise over $n$ for those solutions we examined. Note that we appended a further value to our original $v$ vector when increasing $n$. This generalisation over $n$ is significant as it suggests that optimisation over small problem instances (i.e. problems that we can classically simulate) may lead to improvements for larger systems, in which classical simulation is no longer possible. 


\subsubsection*{Generalisation over v}
\begin{figure}
    \centering
    \includegraphics[width=0.7\textwidth]{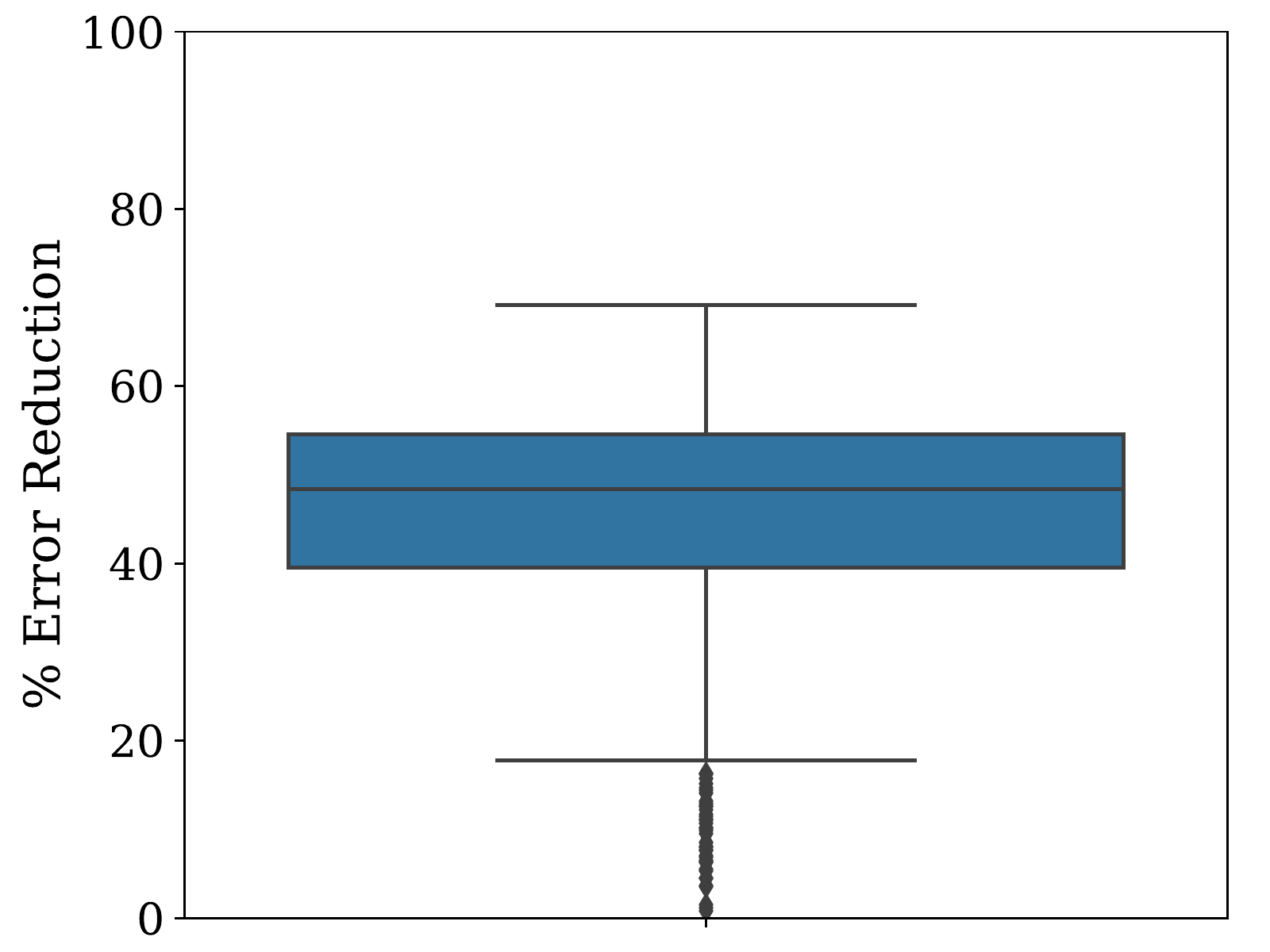}
    \caption{Generalisation over $v$. A $p$-vector optimised for a given Heisenberg chain shows similar error reduction when applied to the simulation of a different chain.}
    \label{fig:generalisation_v}
\end{figure}

Do vectors optimised for specific $v$ values also achieve comparable error reductions for other $v$ (but the same $n$)? We find this to be the case. Taking $30$ $v$ vectors, obtaining an optimised individual for each one, and evaluating the optimised results on the other $29$ $v$ vectors gives rise to Figure \ref{fig:generalisation_v}, which displays a boxplot of the percentage error achieved on unseen $v$. We see that in most cases the optimised $p$ vectors continue to outperform the original Suzuki vector. Thus the results of a single search may be sufficient to optimise many problem instances.

\subsubsection*{Generalisation over r}
\begin{figure}
    \centering
    \includegraphics[width=0.9\textwidth]{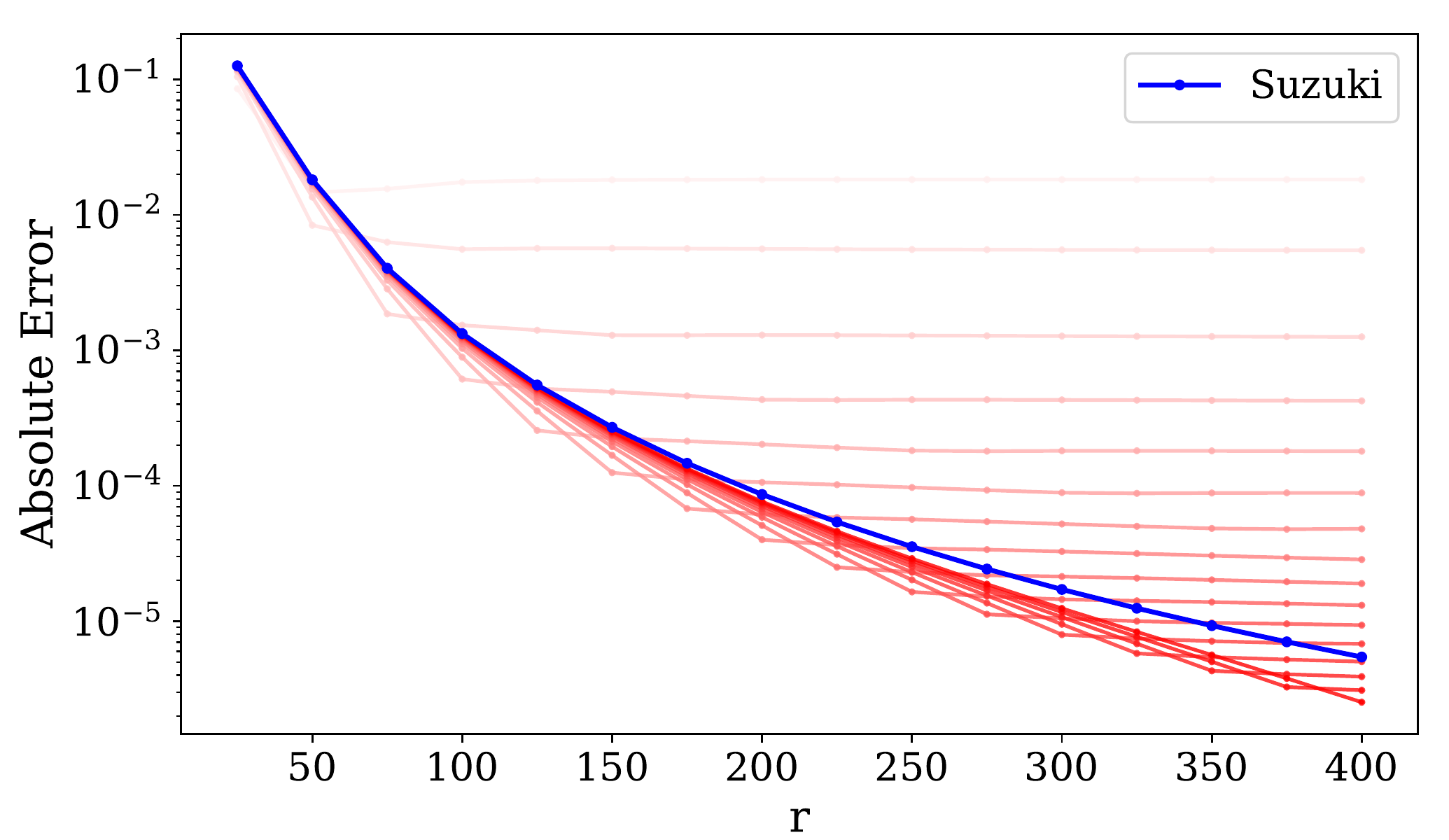}
    \caption{Generalisation over $r$. $p$-vectors optimised for a given $r$ value generalise well to smaller $r$ values, but not to larger ones. This suggests the need for an optimisation approach that encourages generalisation.}
    \label{fig:generalisation_r}
\end{figure}

We also consider how our optimised $p$-vector individuals perform at different $r$. Figure \ref{fig:generalisation_r} shows how $p$- vectors optimised at a specific $r$ fare when evaluated at different $r$. We find that optimised $p$-vectors perform well at values of $r$ less than the $r$ they were optimised against, but in general do not generalise to higher $r$. Note that the apparent divergence for large $r$ on optimised vectors does not contradict the theoretical upper bound in Equation \ref{eq:time_slice}, as our optimised individuals may no longer sum to one. It may be desirable to develop an optimisation method that provides generality over $r$ (see Section \ref{sec:concfw}).

\section{Related Work}
The realisation that a quantum computer would be well positioned to simulate physical systems was originally made in the 1980s by Feynman  \cite{feynman1982simulating} and others, partly motivating the development of quantum computing.
In 1996, Lloyd \cite{lloyd1996} suggested a time-slicing method could indeed achieve efficient simulation for certain Hamiltonians. The theoretical foundations of Suzuki decompositions have been outlined by Suzuki and others \cite{suzuki1991, suzuki1990, hatano2005finding, ruth1983canonical,dhand2014}. This near-term application of quantum computers has lead to much interest and research into how best to perform efficient simulation, particularly for local or sparse Hamiltonians \cite{berry2007efficient, berry2017exponential, wiebe2011simulating}.

As a somewhat counter-intuitive and often multi-objective task, the development of quantum software is an ideal target for metaheuristics and related optimisation methods. To date, most work in this area has focused on quantum circuit synthesis, usually employing Genetic Programming as a search algorithm to rediscover or extend previously known circuits  \cite{spector2004automatic, Massey2005, Stepney2008}, often considering the `quantum cost' or gate count of a solution. The goal is usually to create a circuit of quantum gates that perform the same transformation as given by an input matrix, or else to match measurement outcomes with a given probability. Little work has been done analysing the properties of quantum landscapes, with the exception of Leir and Banzhaf \cite{Leier2003}. Our work is novel in that it empirically optimises a circuit initially derived by a theoretical approach. Perhaps the most closely related work in evolutionary computation is the work by Las Heras \cite{las2016} et al., which considers Heisenberg models as a target for quantum circuit synthesis using a genetic algorithm.



\section{Conclusions and Further Work}
\label{sec:concfw}

We have presented an exposition into the Trotter-Suzuki formulas and quantum simulation for the search community, encoding the problem of error minimisation for simulation as the optimisation of a real-valued vector. We find that a Trotter-Suzuki decomposition does not necessarily represent the optimum for a specific system, and CMA-ES can find large error reductions of around $60 \%$. We find our approach is robust over runs and different problem instances. That Trotter-Suzuki does not represent the optimal simulation of a system is an interesting result: due to the somewhat opaque and unintuitive manner that Trotter-Suzuki exploits the properties of quantum systems, this result was not obvious a priori. 

Whilst our results do not generalise over $r$ and $t$, they \emph{do} generalise over $n$. As the cost of optimization scales with the size of the input system, this offers the ability to optimise on a small system and use the results for larger simulations.

There is potential for further optimisation. In this paper, we tuned the parameters of the Suzuki recurrence relation; a more general approach would be to search the space of recurrence relations using Genetic Programming (GP). Suzuki \cite{suzuki1991} provided two theorems dictating the conditions for cancellation at a given order, and the solution in the recurrence relation used in this paper is only one way of satisfying these constraints. Thus GP could be applied to generate $p$-vector values i.e. evolve Equation \ref{eq:p_vals}, or to search for an alternative recursion relation i.e. to evolve Equation \ref{eq:suzuki_recurrence}. Such an approach may lead to more general results that apply across other problems. A comprehensive exposition examining the numerical performance of the general Suzuki constraints could improve our understanding of how theoretical results translate into practice, and how much room for improvement may be available when super-classical quantum machines arrive.

The ordering of the $e^{H_l}$ terms in Equation \ref{eq:suzuki_second_order}, the second order Suzuki approximation, affects simulation error. We use a specific ordering in this paper, which is justified and discussed in Appendix \ref{app:perm}. Although our ordering produces good results, it is an open question as to whether other permutations may prove superior: search could be used to find useful better permutations.

More generally, the synthesis, refinement, and optimisation of quantum circuits is a large area of optimisation that hitherto remains little-exploited. Most situations involve counter-intuitive solution spaces and often conflicting non-functional constraints: we encourage other researchers to apply search to these problems.

\appendix

\section{Efficient Exponentiation}
\label{app:efficient_exponentiation}
The matrix exponentiation required by the Suzuki decomposition, for example $e^{X_1}$, is expensive as the matrices involves are of size $2^n$. In certain cases, we can invoke the following formula to render this calculation more tractable:

\begin{equation}
e^{A \otimes I + B \otimes I} = e^A \otimes e^B
\end{equation}

\noindent Setting $B=0$ yields

\begin{equation}
e^{A \otimes I} = e^A \otimes I
\end{equation}

\noindent In our case, this results in identities such as the following

\begin{equation}
e^{\alpha X_1 X_2} = e^{\alpha X \otimes X} \otimes I^{n-2}
\end{equation}

\noindent However this reasoning fails for the `wrap around' case $e^{X_n X_1}$

\section{Term Permutations}
\label{sec:perm}
\label{app:perm}

\begin{figure*}
\centering
\includegraphics[width = \textwidth]{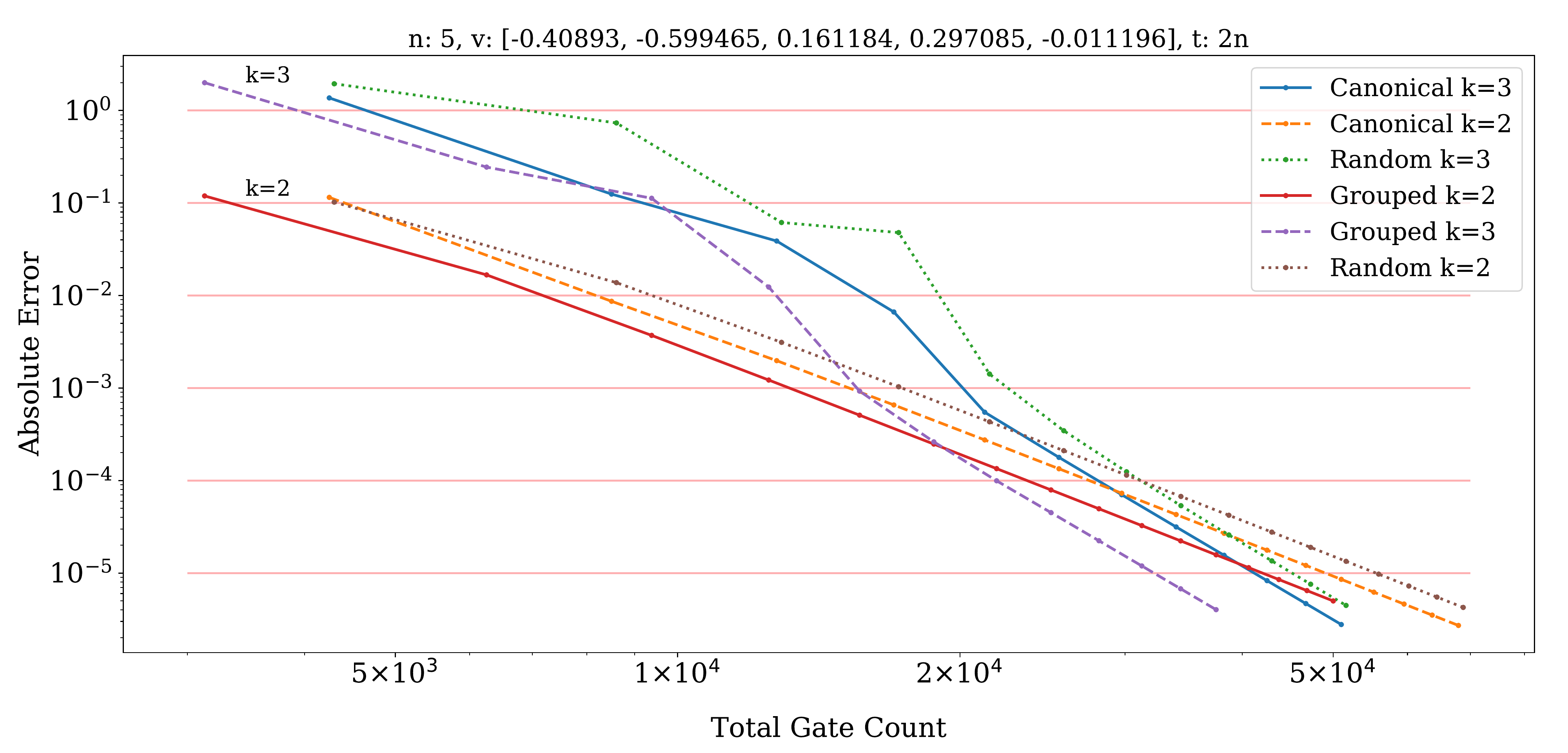}
\caption{Trade-offs achieved with different permutations. The balance between gate count and error is dependent on the ordering of $e^{H_l}$ terms within the decomposition. Here, grouping like terms is shown to outperform the canonical ordering given in Equation \ref{hchain}, and an average over 20 random permutations.}\label{perms}
\end{figure*}

The ordering of the $H_j$ terms in the Hamiltonian sum is arbitrary, as matrix addition is commutative: we may choose alternative orderings of terms in the decomposition, as previously  noted \cite{childs2018}. 

Consider a single permutation of terms that is alternated with its reverse over instances of $S_2$. Both the error and gate count depend on the permutation chosen. To see this, observe that in the total decomposition similar terms will appear multiple times, but perhaps with different phases. If these terms can be brought next to each other via commutativity, then the order of decomposition may remain unchanged but we can collapse multiple terms into a single term by combining their phases. For example,

\begin{equation}
e^{\alpha X_1} e^{\beta X_1} = e^{(\alpha + \beta) X_1}    
\end{equation}

Thus in choosing a permutation we must take into account both error and gate count. 


One way to reduce the number of terms is to group together all $X$ terms (such as $e^{\alpha X_1 X_2}$), $Y$ terms and $Z$ terms, as they commute within these groups. Denoting $\mathbf{X}$ as the product of all $X$ terms (order of which does not matter, as all $X$ terms commute with one another), and similarly for $Y$ and $Z$, we see that repeating $S_2$ $M$ times gives:

\begin{equation}
    \overbrace{ \bigg ( \underset{\substack{\\ n}}{\mathbf{X} \phantom{\bigg )}}  \underset{\substack{ \\ n}}{\mathbf{Y} \phantom{\bigg )}} \underset{\substack{\\ 2n}}{\mathbf{Z}  \hspace{4pt} \mathbf{Z} \phantom{\bigg )}}  \underset{\substack{\\  n}}{\mathbf{Y} \phantom{\bigg )}}  \underset{\substack{ \\ n}}{\mathbf{X}  \bigg ) \bigg ( \mathbf{X}} \phantom{\bigg (}   \underset{\substack{ \\ n}}{\mathbf{Y} \phantom{\bigg )}}  \underset{\substack{\\ 2n}}{\mathbf{Z}  \hspace{4pt} \mathbf{Z} \phantom{\bigg )}} \underset{\substack{\\  n}}{\mathbf{Y} \phantom{\bigg )}}  \underset{\substack{ \\ n}}{\mathbf{X}  \bigg ) \bigg ( \mathbf{X}} \phantom{\bigg (}   \underset{\substack{ \\ n}}{\mathbf{Y} \phantom{\bigg )}}  \underset{\substack{\\ 2n}}{\mathbf{Z}  \hspace{4pt} \mathbf{Z} \phantom{\bigg )}}  \underset{\substack{\\  n}}{\mathbf{Y} \phantom{\bigg )}}  \underset{\substack{ \\ n}}{\mathbf{X}  \bigg ) \ldots} }^{\text{M times}}
\end{equation}

This yields a total of $6n + 5n(M-1) = (5M+1)n$ terms, improving on $8 M n$, the number of gates without considering term cancellations.

We demonstrate this trade-off space in Figure \ref{perms}, where `Grouped' refers to grouping together similar terms (as discussed above), `Canonical' refers to the  permutation arising from Equation \eqref{hchain}, and `Random' is calculated via averaging the errors and gate counts at a given $r$ and $k$ for $20$ random permutations. Whilst grouping together similar terms can increase error (compared with other permutations) at a given order of decomposition, the gate count reductions pareto-dominate other permutation choices, although we do not prove global optimality.

\nocite{*}
\bibliographystyle{plain}
\bibliography{bib}

\begin{thebibliography}{10}

\bibitem{babbush2015}
Ryan Babbush, Jarrod McClean, Dave Wecker, Al{\'a}n Aspuru-Guzik, and Nathan
  Wiebe.
\newblock Chemical basis of trotter-suzuki errors in quantum chemistry
  simulation.
\newblock {\em Physical Review A}, 91(2):022311, 2015.

\bibitem{berry2007efficient}
Dominic~W Berry, Graeme Ahokas, Richard Cleve, and Barry~C Sanders.
\newblock Efficient quantum algorithms for simulating sparse hamiltonians.
\newblock {\em Communications in Mathematical Physics}, 270(2):359--371, 2007.

\bibitem{berry2017exponential}
Dominic~W Berry, Andrew~M Childs, Richard Cleve, Robin Kothari, and Rolando~D
  Somma.
\newblock Exponential improvement in precision for simulating sparse
  hamiltonians.
\newblock In {\em Forum of Mathematics, Sigma}, volume~5. Cambridge University
  Press, 2017.

\bibitem{brown2010}
Katherine~L Brown, William~J Munro, and Vivien~M Kendon.
\newblock Using quantum computers for quantum simulation.
\newblock {\em Entropy}, 12(11):2268--2307, 2010.

\bibitem{campbell2017}
Earl Campbell.
\newblock Shorter gate sequences for quantum computing by mixing unitaries.
\newblock {\em Physical Review A}, 95(4):042306, 2017.

\bibitem{campbell2018random}
Earl Campbell.
\newblock A random compiler for fast hamiltonian simulation.
\newblock {\em arXiv preprint arXiv:1811.08017}, 2018.

\bibitem{childs2017}
Andrew~M. Childs, Dmitri Maslov, Yunseong Nam, Neil~J. Ross, and Yuan Su.
\newblock Toward the first quantum simulation with quantum speedup.
\newblock {\em Proceedings of the National Academy of Sciences},
  115(38):9456--9461, 2018.

\bibitem{childs2018}
Andrew~M Childs, Aaron Ostrander, and Yuan Su.
\newblock Faster quantum simulation by randomization.
\newblock {\em arXiv preprint arXiv:1805.08385}, 2018.

\bibitem{dhand2014}
Ish Dhand and Barry~C Sanders.
\newblock Stability of the trotter--suzuki decomposition.
\newblock {\em Journal of Physics A: Mathematical and Theoretical},
  47(26):265206, 2014.

\bibitem{feynman1982simulating}
Richard~P Feynman.
\newblock Simulating physics with computers.
\newblock {\em International journal of theoretical physics}, 21(6-7):467--488,
  1982.

\bibitem{DEAP}
F{\'e}lix-Antoine Fortin, Fran{\c{c}}ois-Michel~De Rainville, Marc-Andr{\'e}
  Gardner, Marc Parizeau, and Christian Gagn{\'e}.
\newblock Deap: Evolutionary algorithms made easy.
\newblock {\em Journal of Machine Learning Research}, 13(Jul):2171--2175, 2012.

\bibitem{georgescu2014}
I.~M. Georgescu, S.~Ashhab, and Franco Nori.
\newblock Quantum simulation.
\newblock {\em Rev. Mod. Phys.}, 86:153--185, Mar 2014.

\bibitem{Hansen2003}
Nikolaus Hansen, Sibylle~D M{\"u}ller, and Petros Koumoutsakos.
\newblock Reducing the time complexity of the derandomized evolution strategy
  with covariance matrix adaptation (cma-es).
\newblock {\em Evolutionary computation}, 11(1):1--18, 2003.

\bibitem{Hansen2001}
Nikolaus Hansen and Andreas Ostermeier.
\newblock Completely derandomized self-adaptation in evolution strategies.
\newblock {\em Evolutionary computation}, 9(2):159--195, 2001.

\bibitem{hatano2005finding}
Naomichi Hatano and Masuo Suzuki.
\newblock Finding exponential product formulas of higher orders.
\newblock In {\em Quantum annealing and other optimization methods}, pages
  37--68. Springer, 2005.

\bibitem{hempel2018quantum}
Cornelius Hempel, Christine Maier, Jonathan Romero, Jarrod McClean, Thomas
  Monz, Heng Shen, Petar Jurcevic, Ben Lanyon, Peter Love, Ryan Babbush, et~al.
\newblock Quantum chemistry calculations on a trapped-ion quantum simulator.
\newblock {\em arXiv preprint arXiv:1803.10238}, 2018.

\bibitem{jones2018quest}
Tyson Jones, Anna Brown, Ian Bush, and Simon Benjamin.
\newblock Quest and high performance simulation of quantum computers.
\newblock {\em arXiv preprint arXiv:1802.08032}, 2018.

\bibitem{las2016}
Urtzi Las~Heras, Unai Alvarez-Rodriguez, Enrique Solano, and Mikel Sanz.
\newblock Genetic algorithms for digital quantum simulations.
\newblock {\em Physical review letters}, 116(23):230504, 2016.

\bibitem{Leier2003}
Andr{\'e} Leier and Wolfgang Banzhaf.
\newblock Evolving hogg’s quantum algorithm using linear-tree gp.
\newblock In {\em Genetic and Evolutionary Computation Conference}, pages
  390--400. Springer, 2003.

\bibitem{lloyd1996}
Seth Lloyd.
\newblock Universal quantum simulators.
\newblock {\em Science}, pages 1073--1078, 1996.

\bibitem{luitz2015many}
David~J Luitz, Nicolas Laflorencie, and Fabien Alet.
\newblock Many-body localization edge in the random-field heisenberg chain.
\newblock {\em Physical Review B}, 91(8):081103, 2015.

\bibitem{Massey2005}
Paul Massey, John~A Clark, and Susan Stepney.
\newblock Evolution of a human-competitive quantum fourier transform algorithm
  using genetic programming.
\newblock In {\em Proceedings of the 7th annual conference on Genetic and
  evolutionary computation}, pages 1657--1663. ACM, 2005.

\bibitem{mcardle2018}
Sam McArdle, Suguru Endo, Alan Aspuru-Guzik, Simon Benjamin, and Xiao Yuan.
\newblock Quantum computational chemistry.
\newblock {\em arXiv preprint arXiv:1808.10402}, 2018.

\bibitem{nandkishore2015many}
Rahul Nandkishore and David~A Huse.
\newblock Many-body localization and thermalization in quantum statistical
  mechanics.
\newblock {\em Annu. Rev. Condens. Matter Phys.}, 6(1):15--38, 2015.

\bibitem{nielsen2002quantum}
Michael~A Nielsen and Isaac Chuang.
\newblock Quantum computation and quantum information, 2002.

\bibitem{oliphant2006guide}
Travis~E Oliphant.
\newblock {\em A guide to NumPy}, volume~1.
\newblock Trelgol Publishing USA, 2006.

\bibitem{pal2010many}
Arijeet Pal and David~A Huse.
\newblock Many-body localization phase transition.
\newblock {\em Physical review b}, 82(17):174411, 2010.

\bibitem{preskill2018}
John Preskill.
\newblock Quantum {C}omputing in the {NISQ} era and beyond.
\newblock {\em {Quantum}}, 2:79, August 2018.

\bibitem{rieffel2000introduction}
Eleanor Rieffel and Wolfgang Polak.
\newblock An introduction to quantum computing for non-physicists.
\newblock {\em ACM Computing Surveys (CSUR)}, 32(3):300--335, 2000.

\bibitem{ruth1983canonical}
Ronald~D Ruth.
\newblock A canonical integration technique.
\newblock {\em IEEE Trans. Nucl. Sci.}, 30(CERN-LEP-TH-83-14):2669--2671, 1983.

\bibitem{shende2006}
V.~V. Shende, S.~S. Bullock, and I.~L. Markov.
\newblock Synthesis of quantum-logic circuits.
\newblock {\em IEEE Transactions on Computer-Aided Design of Integrated
  Circuits and Systems}, 25(6):1000--1010, June 2006.

\bibitem{somma2016}
Rolando~D Somma.
\newblock A trotter-suzuki approximation for lie groups with applications to
  hamiltonian simulation.
\newblock {\em Journal of Mathematical Physics}, 57(6):062202, 2016.

\bibitem{spector2004automatic}
Lee Spector.
\newblock {\em Automatic Quantum Computer Programming: a genetic programming
  approach}, volume~7.
\newblock Springer Science \& Business Media, 2004.

\bibitem{Stepney2008}
Susan Stepney and John~A Clark.
\newblock Searching for quantum programs and quantum protocols.
\newblock {\em Journal of Computational and Theoretical Nanoscience},
  5(5):942--969, 2008.

\bibitem{suzuki1990}
Masuo Suzuki.
\newblock Fractal decomposition of exponential operators with applications to
  many-body theories and monte carlo simulations.
\newblock {\em Physics Letters A}, 146(6):319--323, 1990.

\bibitem{suzuki1991}
Masuo Suzuki.
\newblock General theory of fractal path integrals with applications to
  many-body theories and statistical physics.
\newblock {\em Journal of Mathematical Physics}, 32(2):400--407, 1991.

\bibitem{wiebe2011simulating}
Nathan Wiebe, Dominic~W Berry, Peter H{\o}yer, and Barry~C Sanders.
\newblock Simulating quantum dynamics on a quantum computer.
\newblock {\em Journal of Physics A: Mathematical and Theoretical},
  44(44):445308, 2011.

\bibitem{zhang2019}
Yikang Zhang and Thomas Barthel.
\newblock Optimized higher-order lie-trotter-suzuki decompositions for two and
  more terms.
\newblock {\em Bulletin of the American Physical Society}, 2019.

\end{thebibliography}

\end{document}